\documentclass[runningheads]{llncs}

\usepackage[T1]{fontenc}
\usepackage{graphicx}

\usepackage{amsmath}
\usepackage{amssymb}
\usepackage{bm}
\usepackage{booktabs}
\usepackage{multirow}
\usepackage{array}
\usepackage{xcolor}
\usepackage{url}
\usepackage{hyperref}
\usepackage{cite}
\usepackage{todonotes}

\usepackage{color}

\newcommand{\R}{\mathbb{R}}
\newcommand{\Sig}{\boldsymbol{\Sigma}}
\newcommand{\Lb}{\mathbf{L}}
\newcommand{\cb}{\mathbf{c}}
\newcommand{\pb}{\mathbf{p}}
\newcommand{\fb}{\mathbf{f}}
\newcommand{\vb}{\mathbf{v}}
\newcommand{\T}{\mathbf{T}}

\begin{document}

\title{UQ-Loc: Uncertainty-Aware LiDAR Scene Coordinate Regression
}

\titlerunning{UQ-Loc: Uncertainty-Aware LiDAR SCR}

\author{Jacek Komorowski\inst{1}\orcidID{0000-0001-6906-4318}}
\authorrunning{J. Komorowski}

\institute{Warsaw University of Technology, Warsaw, Poland\\
\email{jacek.komorowski@pw.edu.pl}}

\maketitle

\begin{abstract}
LiDAR-based Scene Coordinate Regression (SCR) maps point clouds directly
to 3D scene coordinates, enabling precise 6-DoF localisation without
explicit map retrieval. However, existing methods produce deterministic
predictions, discarding aleatoric uncertainty that could improve
robustness and downstream decision-making. We present \emph{UQ-Loc},
which extends the LightLoc architecture~\cite{lightloc} with an anisotropic Gaussian covariance head that predicts
a full $3{\times}3$ positive-definite covariance matrix per
voxel. Training uses a Negative Log-Likelihood (NLL) loss augmented with
a kNN-based spatial smoothness regulariser, while inference employs a
modified SC2-PCR solver~\cite{sc2pcr} with uncertainty-weighted seed
scoring and a Mahalanobis-distance inlier test. 
We adopt Expected Calibration Error (ECE) as a principled metric for evaluating the quality of the predicted uncertainty.
Experiments demonstrate that UQ-Loc achieves consistent improvement in 6-DoF localization accuracy while producing well-calibrated covariances.


\end{abstract}

\section{Introduction}
\label{sec:intro}

Accurate and reliable 6-DoF localisation is fundamental to autonomous
driving and mobile robotics. LiDAR sensors provide geometrically rich,
weather-robust point clouds that complement or replace camera-based
systems. Scene Coordinate Regression (SCR), the direct mapping of sensor
observations to 3D scene coordinates, has demonstrated strong
performance in camera-based localisation~\cite{dsacstar} and is
now extending to the LiDAR domain~\cite{lightloc}. 
In SCR, a network maps each voxel in an input scan to its corresponding world-coordinate location, 
and the resulting point-to-world correspondences are passed to a RANSAC-style solver to recover the 6-DoF pose.
A critical limitation of current LiDAR SCR methods is that they produce
purely deterministic predictions: each voxel yields a single
point estimate $\hat{\cb}_i \in \R^3$, with no indication of how much
that estimate should be trusted. This discards valuable information.
Uncertain regions, such as glass building façades, vegetation, dynamic objects,
and structural boundaries, generate poor correspondences that degrade
RANSAC performance when treated identically to high-quality ones. More
broadly, the inability to communicate uncertainty prevents integration
with probabilistic downstream modules such as factor-graph SLAM
backends, occupancy estimation, and failure detection systems.

We address these limitations by presenting \textbf{UQ-Loc}, which
augments LightLoc~\cite{lightloc} with full anisotropic per-voxel uncertainty.
The main contributions of this work are as follows:
\begin{enumerate}
\item \textbf{Anisotropic uncertainty head.} We extend the LightLoc regression head 
with a lightweight branch that regresses a raw six-parameter Cholesky parameterisation per voxel.
We replace the original L1 regression loss with a multivariate Gaussian Negative Log-Likelihood (NLL) that jointly optimises coordinate accuracy and covariance calibration, augmented by a kNN smoothness regulariser.


\item \textbf{Uncertainty-aware SC2-PCR solver.} We modify the SC2-PCR
  solver to modulate spectral seed scores by down-weighting uncertain correspondences, and replace the fixed  Euclidean inlier threshold with a Mahalanobis-distance criterion.


\end{enumerate}


\section{Related Work}
\label{sec:related}

\subsubsection{LiDAR Place Recognition and Localisation.}
Metric LiDAR localisation is commonly achieved by matching local features
or iterative closest-point alignment against a pre-built map,
but such methods are sensitive to initialisation and map density.
Descriptor-based retrieval methods, including PointNetVLAD~\cite{pointnetvlad} and MinkLoc3D~\cite{minkloc3d}, 
aggregate global descriptors for coarse place retrieval but do not directly yield metric 6-DoF poses.
Absolute Pose Regression (APR) methods address this by training a network
to directly regress the 6-DoF pose from a sensor observation~\cite{posenet},
but they generalise poorly to unseen viewpoints and typically yield lower
accuracy than geometry-based alternatives due to the difficulty of
encoding absolute pose as a learnable output.
Scene Coordinate Regression (SCR) methods take a different approach:
rather than regressing the pose directly, the network maps each point in
the input scan to its corresponding location in a pre-defined world
coordinate frame, and the resulting point-to-world correspondences are
passed to a robust solver to recover the pose.
SCR-based methods~\cite{sgloc,lightloc} thus preserve geometric
structure in the intermediate representation, leading to higher accuracy
than APR. We review LiDAR SCR methods below.

\subsubsection{Scene Coordinate Regression (SCR).}
SCR was introduced by Shotton et al.~\cite{shotton2013} for RGB-D
cameras. Subsequent deep learning formulations substantially improved
scalability and accuracy, notably DSAC*~\cite{dsacstar} and
SGLoc~\cite{sgloc} extended the SCR paradigm to outdoor LiDAR by
decoupling pose estimation into a point cloud correspondence regression
stage followed by RANSAC-based pose recovery, encoding scene geometry
implicitly in the network parameters and achieving sub-metre accuracy
on large-scale outdoor benchmarks.
LightLoc~\cite{lightloc} retains the correspondence-regression paradigm
while reducing training time through a classification-conditioned regression head and sparse convolutional
features. 
None of these methods model per-point predictive covariance; UQ-Loc addresses this gap by extending LightLoc with full anisotropic uncertainty estimation

\subsubsection{Uncertainty Estimation in Localisation.}
Kendall and Cipolla~\cite{kendallcipolla} established heteroscedastic
aleatoric uncertainty for camera pose regression; their scalar output
cannot capture directional anisotropy.
Bayesian approaches quantify epistemic uncertainty but are computationally prohibitive~\cite{kendallcipolla}.
In visual SCR, Brachmann and Rother~\cite{brachmann2018} predicted
per-pixel inlier probabilities to guide hypothesis selection. 
This is a related but weaker form of per-point uncertainty than the full covariance we propose.
In 3D object detection, Cholesky-parameterised covariances have been
used to ensure positive-definiteness without projection
steps~\cite{he2020deep}. We adopt the same parameterisation for 3D
scene coordinates.

\subsubsection{Spectral Matching and RANSAC Solvers.}
SC2-PCR~\cite{sc2pcr} estimates rigid transformations by building a
compatibility graph over point correspondences and extracting the leading
eigenvector via power iteration to score each correspondence by its
geometric consistency with the full set; seeds for local hypothesis
generation are then selected by non-maximum suppression on these spectral
scores.
This formulation is particularly well-suited to the SCR setting, where
the number of correspondences is large (one per voxel) and the inlier
ratio can be low in challenging scenes, conditions under which standard
RANSAC requires prohibitively many iterations to find a good hypothesis.
We therefore build on SC2-PCR and extend it with uncertainty-weighted seed scoring and a
Mahalanobis-distance inlier test.

\section{Method}
\label{sec:method}

\subsection{UQ-Loc Overview and Base Architecture}
\label{sec:background}

UQ-Loc extends LightLoc~\cite{lightloc} with three components that add
principled uncertainty estimation while leaving the coordinate regression
backbone intact: \textbf{(i)} an anisotropic uncertainty head
that predicts a full $3{\times}3$ covariance $\Sig_i$ per voxel
(Section~\ref{sec:unchhead}); \textbf{(ii)} a Negative Log-Likelihood
training loss with spatial smoothness that jointly optimises coordinate
accuracy and covariance calibration (Section~\ref{sec:loss}); and
\textbf{(iii)} a modified SC2-PCR solver that uses predicted covariances
for uncertainty-weighted seed scoring and Mahalanobis inlier testing
(Section~\ref{sec:solver}).

The base architecture operates as follows.
Given a LiDAR scan, we extract $N$ voxels and produce a 512-dimensional
feature vector $\fb_i \in \R^{512}$ per voxel via a frozen sparse 3D
CNN backbone (MinkUNet-style~\cite{minkowski}).
A classification head assigns each voxel to one of $K$ place clusters;
the resulting cluster-conditioned features are passed to the regression
head, which in the original LightLoc predicts a single world coordinate
per voxel:
\begin{equation}
  \hat{\cb}_i = \text{Reg\_Head}(\fb_i) \in \R^3,
  \qquad
  \mathcal{L}_{\text{L1}} = \sum_{i=1}^{N} \|\hat{\cb}_i - \cb^*_i\|_1 ,
\end{equation}
where $\cb^*_i$ is the ground-truth world coordinate.
The correspondences $\{(\pb_i, \hat{\cb}_i)\}$, sensor-frame point
$\pb_i$ paired with predicted world coordinate $\hat{\cb}_i$, are
passed to the SC2-PCR solver to recover the 6-DoF pose $\T \in SE(3)$.
UQ-Loc augments this pipeline by additionally predicting $\Sig_i$ per
voxel, so the regression head now returns a tuple $(\hat{\mathbf{c}}_i, \mathbf{v}_i)$, where $\mathbf{v}_i \in \mathbb{R}^6$ parameterises the covariance $\Sigma_i$ via a Cholesky factorisation (§\ref{sec:unchhead}).
The modified components are described in the following
subsections.

\subsection{Anisotropic Uncertainty Head}
\label{sec:unchhead}

We add an \emph{UncertaintyHead} branch inside the existing regression
head, sharing the 512-dimensional residual-block features with the
coordinate branch.
Three fully-connected layers (512\,${\to}$\,256\,${\to}$\,128\,${\to}$\,6)
output a compact 6-dimensional vector $\vb_i \in \R^6$ that
parameterises a lower-triangular Cholesky factor:
\begin{equation}
  \Lb_i = \begin{pmatrix}
    \text{softplus}(v_1) & 0 & 0 \\
    v_2 & \text{softplus}(v_3) & 0 \\
    v_4 & v_5 & \text{softplus}(v_6)
  \end{pmatrix}, \quad
  \Sig_i = \Lb_i \Lb_i^\top \in \R^{3\times 3}.
  \label{eq:chol}
\end{equation}
Applying \texttt{softplus} to the three diagonal entries guarantees
strict positivity ($L_{kk} > 0$) and hence $\Sig_i \succ 0$ by
construction.  A $3{\times}3$ lower-triangular matrix has exactly
$6$ free parameters (3 diagonal $+$ 3 sub-diagonal), matching
the degrees of freedom of a general $3{\times}3$ symmetric
positive-definite matrix.

Compared with simpler alternatives, the scalar isotropic
variant ($\sigma_i^2\mathbf{I}$, 1 DOF) cannot represent directional
anisotropy; the axis-aligned diagonal variant (3 DOF) cannot capture
correlations between coordinate axes.  The full Cholesky
parameterisation (6 DOF) subsumes both and can represent elongated
ellipsoidal uncertainty aligned with scene structure. For example,
uncertainty along the LiDAR ray direction at range discontinuities or
planar surfaces.

The regression head returns a tuple $(\hat{\cb}_i,\, \vb_i)$;
the coordinate branch is structurally unchanged, so the modification
introduces no additional computational cost at inference when
uncertainty estimates are not required.

\subsection{NLL Training Loss with Spatial Smoothness}
\label{sec:loss}

\subsubsection{NLL Loss.}


We model scene coordinate prediction as a heteroscedastic Gaussian: the
true world coordinate $\cb^*_i$ is drawn from
$\cb^*_i \sim \mathcal{N}(\hat{\cb}_i,\, \Sig_i)$, where $\hat{\cb}_i$
is the predicted mean and $\Sig_i \succ 0$ is the predicted covariance.
Under this model, the probability of observing $\cb^*_i$ is
\begin{equation*}
  p(\cb^*_i) = \frac{1}{(2\pi)^{3/2}\det(\Sig_i)^{1/2}}
  \exp\!\left(-\tfrac{1}{2}(\cb^*_i - \hat{\cb}_i)^\top
    \Sig_i^{-1}(\cb^*_i - \hat{\cb}_i)\right).
\end{equation*}
Maximising the log-likelihood over the training set is equivalent to
minimising the negative log-likelihood (NLL)~\cite{kendallcipolla},
which for a single voxel gives:
\begin{equation}
  \ell_i = \frac{1}{2}\left[
      \underbrace{\|\Lb_i^{-1}(\cb^*_i - \hat{\cb}_i)\|^2}_{\text{Mahalanobis term}}
      + \underbrace{\log\det(\Sig_i)}_{\text{log-det penalty}}
    \right]
  = \frac{1}{2}\left[
      \|\Lb_i^{-1}(\cb^*_i - \hat{\cb}_i)\|^2
      + 2\sum_{k=1}^{3}\log (L_i)_{kk}
    \right],
  \label{eq:nll}
\end{equation}
where we drop the constant $\tfrac{3}{2}\log(2\pi)$ as it does not affect
optimisation, and the simplification
$\log\det(\Sig_i) = 2\sum_k \log(L_i)_{kk}$ follows from
$\det(\Sig_i) = \det(\Lb_i\Lb_i^\top) = \det(\Lb_i)^2 = \prod_k L_{kk}^2$.

The two terms play opposing roles and jointly prevent degenerate
solutions.
The \textbf{Mahalanobis term} $\|\Lb_i^{-1}(\cb^*_i -
\hat{\cb}_i)\|^2$ is the squared prediction error normalised by
the predicted uncertainty: a voxel with large covariance $\Sig_i$
contributes a small residual even if the raw error $\|\cb^*_i -
\hat{\cb}_i\|$ is large. This gives the network an incentive to assign
high uncertainty to hard-to-predict voxels (e.g.\ dynamic objects,
glass) — effectively down-weighting their contribution to the gradient
signal, a property Kendall and Gal~\cite{kendall2017uncertainties} term
\emph{learned loss attenuation}.
The \textbf{log-det penalty} $\log\det(\Sig_i)$ counteracts this:
inflating $\Sig_i$ to minimise the Mahalanobis term increases the
log-det, penalising the network for predicting uniformly large
covariances. 
Compared to the scalar 1-D formulation of~\cite{kendallcipolla}, our
loss operates on a full $3{\times}3$ covariance, capturing anisotropic
uncertainty (e.g.\ larger variance along the LiDAR ray direction than
perpendicular to it).

The Mahalanobis term $\|\Lb_i^{-1}(\cb^*_i - \hat{\cb}_i)\|^2$ is
computed via a triangular forward-substitution solve, which is
numerically stable and avoids forming $\Sig_i^{-1}$ explicitly.

\subsubsection{Spatial Smoothness Regulariser.}
Raw NLL optimisation can produce spatially noisy covariance fields.
Neighbouring voxels observing the same surface should carry similar
uncertainty. We penalise the Frobenius-norm difference between adjacent
covariances:
\begin{equation}
  \mathcal{L}_{\text{smooth}} = \lambda \sum_{i}\sum_{j \in \mathcal{N}_k(i)}
    \|\Sig_i - \Sig_j\|_F^2 ,
  \label{eq:smooth}
\end{equation}
where $\mathcal{N}_k(i)$ is the set of $k{=}8$ nearest neighbours of
voxel $i$ in Euclidean space. 
To reduce the complexity, the sum is capped at 2048 randomly sampled pairs per batch. 
We set $\lambda{=}0.001$; the total loss is $\mathcal{L} = \mathcal{L}_{\text{NLL}} + \mathcal{L}_{\text{smooth}}$.

The Relative Scene Difficulty (RSD) hard-negative mining from
LightLoc~\cite{lightloc}, which samples scans whose median loss exceeds
the running 75th-percentile threshold, is preserved, with the per-scan
median NLL replacing the median L1 error.

\subsection{Uncertainty-Aware SC2-PCR Solver}
\label{sec:solver}

\subsubsection{Uncertainty-Weighted Seed Scoring.}
SC2-PCR assigns a spectral compatibility score $s_i$ to each
correspondence. High-uncertainty voxels are less reliable as RANSAC
seeds; we modulate the score by an uncertainty-volume penalty:
\begin{equation}
  \tilde{s}_i = \frac{s_i}{1 + \alpha\cdot\det(\Sig_i)^{1/3}} ,
  \label{eq:seedscore}
\end{equation}
where $\det(\Sig_i)^{1/3}$ is the geometric mean of the three principal
variances, and $\alpha{=}1$ in all experiments.

\subsubsection{Mahalanobis Inlier Test.}
The standard Euclidean inlier criterion $\|\T\pb_i - \hat{\cb}_i\|_2 <
\tau$ treats all correspondences symmetrically. We replace it with an
anisotropic test:
\begin{equation}
  d_M^2(i,\T)
  = (\T\pb_i - \hat{\cb}_i)^\top \Sig_i^{-1} (\T\pb_i - \hat{\cb}_i)
  < \tau .
  \label{eq:mahal}
\end{equation}
The threshold $\tau = 4$, chosen experimentally, is approximately the 74th percentile of $\chi^2_3$.
Under perfect calibration, 74\% of inliers at the true pose will satisfy the test.
Computationally, $\Sig_i^{-1}$ is never formed explicitly; instead
we solve $\Lb_i \mathbf{z} = (\T\pb_i - \hat{\cb}_i)$ via
\texttt{solve\_triangular} and compute $d_M^2 = \|\mathbf{z}\|^2$.
The solve is applied in a single batched GPU call across all seeds and
correspondences simultaneously.


\section{Experiments}
\label{sec:experiments}

\subsection{Datasets and Metrics}
\label{sec:datasets}

We evaluate UQ-Loc on two large-scale outdoor LiDAR localisation benchmarks: Oxford RobotCar and NCLT.

\textbf{Oxford RobotCar}~\cite{oxfordrobotcar} is collected by sensors
mounted on an Nissan LEAF platform, designed for
urban scene localisation tasks. Each trajectory spans approximately
10\,km, covering an area of about 2\,km$^2$. Point clouds are generated
by dual Velodyne HDL-32E LiDAR sensors, with ground-truth poses provided
by a GPS/INS system. The dataset is captured under varied weather and
traffic conditions, making it highly suitable for comprehensive method
evaluation. 
We employ \textbf{QEOxford}, a quality-enhanced version of the Oxford RobotCar dataset.
It minimises GPS/INS errors through trajectory alignment techniques, 
yielding more accurate ground-truth poses that have been shown to benefit localisation training~\cite{sgloc}.
We use the same training and test sequences as LightLoc~\cite{lightloc}
with a voxel size of 0.25\,m and 25 place clusters.

\textbf{NCLT}~\cite{nclt} is collected by sensors mounted on a Segway robotic
platform on the University of Michigan's North Campus.
Each trajectory spans approximately 5.5\,km and covers an area of about
0.45\,km$^2$.
Point clouds are acquired by a Velodyne HDL-32E LiDAR sensor, with
ground-truth poses provided by a SLAM-based post-processing pipeline.
The dataset consists of 27 sessions recorded approximately bi-weekly over
15 months, encompassing a diverse range of
conditions including seasonal changes, varying illumination, and both
indoor and outdoor environments, making it particularly challenging for
long-term localisation.
We use the same train and test sequences as LightLoc~\cite{lightloc} with a voxel size of 0.3\,m and 100 place clusters.

\textbf{Metrics.} We report mean and median translation error~(m),
mean and median rotation error~(\textdegree), recall at
$(0.5\,\text{m},\,1^{\circ})$, $(1\,\text{m},\,2^{\circ})$, and $(2\,\text{m},\,5^{\circ})$
thresholds, and Expected Calibration Error (ECE) calculated using formula~\ref{eq:ece}.

\subsection{Implementation Details}

The MinkUNet backbone and classification heads are frozen from LightLoc pre-training.
The UncertaintyHead uses ReLU activations in fc\_chol1/2
and no activation in fc\_chol3 before the \texttt{chol\_to\_matrix}
mapping. Training uses Adam ($\text{lr}{=}10^{-3}$, cosine decay, 250
epochs). Smoothness parameters:
$k{=}8$, cap\,=\,2048, $\lambda{=}0.001$. SC2-PCR parameters: $\alpha{=}1.0$,
$\chi^2$ threshold 4. All experiments run on a single NVIDIA A100 GPU.

\subsection{Comparison with State of the Art}

Table~\ref{tab:quant-lightloc} compares UQ-Loc against our LightLoc reproduction on both the QEOxford and NCLT datasets.
UQ-Loc consistently outperforms the baseline across all metrics.
On QEOxford, the mean translation error is reduced by 16\% and the mean rotation error by 35\%. 
The median translation and rotation errors decrease by 24\% and 33\%, respectively.
The improvement is most pronounced in the recall metric at the strict $(0.5\,\text{m},\,1^{\circ})$ threshold. 
UQ-Loc achieves 49.5\% recall compared to 26.5\% for LightLoc, a relative gain
of 87\%, confirming that it recovers a substantially larger fraction of high-precision poses.

On NCLT, UQ-Loc achieves similar gains: reduction of mean translation and rotation errors by 18\% and 27\%;
median translation and rotation errors by 43\% and 19\%, respectively.

\begin{table}
\caption{Comparison with the baseline LightLoc method.
We report the mean and median translation (TE) and rotation (RE) errors, alongside recall at various translation and rotation error thresholds.
$\ddagger$ denotes our reproduction using weights available on GitHub.}
\label{tab:quant-lightloc}
\centering
\setlength{\tabcolsep}{5pt}
\begin{tabular}{lccccccc}
\toprule
\multirow{2}{*}{Method}
 & \multicolumn{2}{c}{Mean$\downarrow$}
 & \multicolumn{2}{c}{Median$\downarrow$}
 & \multicolumn{3}{c}{Recall@\emph{threshold}$\uparrow$}
\\
\cmidrule(lr){2-3}\cmidrule(lr){4-5}\cmidrule(lr){6-8}
& TE\,(m) & RE\,(\textdegree) 
& TE\,(m) & RE\,(\textdegree)
& 0.5m,1\textdegree & 1m,2\textdegree & 2m,5\textdegree \\
\midrule
\multicolumn{8}{c}{QEOxford dataset (average over four test sequences)} \\
LightLoc$\ddagger$~\cite{lightloc} & 0.81 & 1.08 & 0.59 & 0.90 & 0.265 & 0.819 & 0.965 \\
\textbf{UQ-Loc (ours)}
& \textbf{0.68}	& \textbf{0.70} & \textbf{0.45} & \textbf{0.60} & \textbf{0.495} & \textbf{0.892} &  \textbf{0.976} \\
\midrule
\multicolumn{8}{c}{NCLT dataset (average over four test sequences)} \\
LightLoc$\ddagger$~\cite{lightloc} &  1.65 & 2.85 & 0.58 & 1.91 & 0.095 & 0.467 & 0.883 \\
\textbf{UQ-Loc (ours)}
& \textbf{1.35} & \textbf{2.09} & \textbf{0.33} & \textbf{1.55} & \textbf{0.207} & \textbf{0.636} & \textbf{0.943} \\
\bottomrule
\end{tabular}
\end{table}

Table~\ref{tab:sota-qeoxford} compares UQ-Loc against published APR and SCR baselines on the QEOxford dataset.
UQ-Loc achieves the best translation and rotation accuracy across all test sequences.
Averaged over all sequences, it reduces the mean translation estimation error by 16\% (0.68\,m versus 0.81\,m) and the mean rotation estimation error by 20\% ($0.70^{\circ}$ versus $0.87^{\circ}$) compared to the best previously published results.
The gain is consistent across all sequences, confirming that the Mahalanobis inlier test in the modified
SC2-PCR solver generalises beyond a single recording condition.

\begin{table}
\caption{Comparison with SOTA on the test sequences from the QEOxford dataset. We report the mean translation (TE) and rotation (RE) errors. Best results in \textbf{bold}, second best \underline{underlined}. 
$\dagger$~denotes results from the LightLoc paper~\cite{lightloc}, $\ddagger$~our reproduction using weights available on GitHub.
}
\label{tab:sota-qeoxford}
\centering
\setlength{\tabcolsep}{3pt}
\begin{tabular}{clccccc}
\toprule
 & \multirow{2}{*}{Method}
  & \multicolumn{5}{c}{Mean TE\,(m)/RE\,(\textdegree)}
\\
\cmidrule(lr){3-7}
& & 01-15-13-06 
& 01-17-13-26 
& 01-17-14-03 
& 01-18-14-14 
  & Average \\
\midrule
\multirow{4}{*}{\rotatebox{90}{APR}} 
& PointLoc$\dagger$~\cite{wang2021pointloc}
& 10.75/2.36 & 11.07/2.21 & 11.53/1.92 & 9.82/2.07 & 10.79/2.14  \\
& PoseMinkLoc$\dagger$~\cite{yu2022lidar}
& 6.77/1.84 & 8.84/1.84 & 8.08/1.69 & 6.56/2.06 & 7.56/1.86  \\
& PoseSOE$\dagger$~\cite{yu2022lidar}
& 4.17/1.76 & 6.16/1.81 & 5.42/1.87 & 4.16/1.70 & 4.98/1.79  \\
& DiffLoc$\dagger$~\cite{li2024diffloc}
& 2.03/\underline{1.04} & 1.78/\underline{0.79} & 2.05/\underline{0.83} & 1.56/\underline{0.83} & 1.86/\underline{0.87}  \\
\midrule
\multirow{5}{*}{\rotatebox{90}{SCR}} 
& SGLoc$\dagger$~\cite{sgloc}
& 1.79/1.67 & 1.81/1.76 & 1.33/1.59 & 1.19/1.39 & 1.53/1.60 \\
& LightLoc$\dagger$~\cite{lightloc}
& \underline{0.82}/1.12 &  \underline{0.85}/1.07 & 0.81/1.11 & \underline{0.82}/1.16 & 0.83/1.12  \\
& LightLoc$\ddagger$~\cite{lightloc}
& 0.84/1.07 &  0.87/1.06 & \underline{0.71}/1.06 & 0.84/1.13 & \underline{0.81}/1.08  \\
& \textbf{UQ-Loc (ours)}
& \textbf{0.71}/\textbf{0.70} & \textbf{0.71}/\textbf{0.66} & \textbf{0.56}/\textbf{0.66} & \textbf{0.72}/\textbf{0.77} & \textbf{0.68}/\textbf{0.70} \\
\bottomrule
\end{tabular}
\end{table}

\begin{figure}
    \centering
    \includegraphics[width=0.5\linewidth]{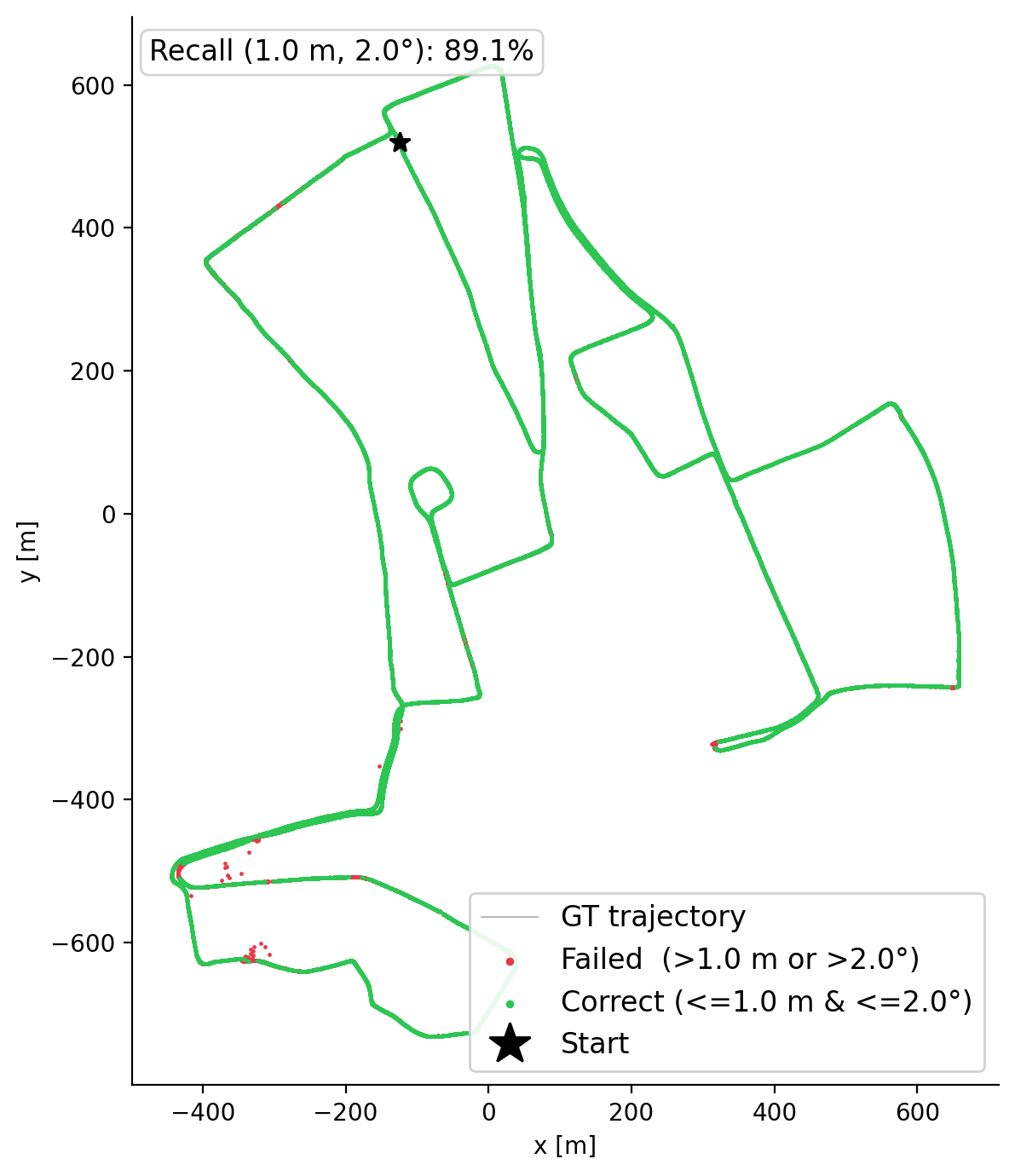}
    \caption{Predicted trajectory for the \texttt{2019-01-17} QEOxford test sequence.
    Green dots indicate correctly localised scans
    (translation error $\leq 1$\,m and rotation error $\leq 2^\circ$);
    red dots indicate failures.}
    \label{fig:trajectory}
\end{figure}

Figure~\ref{fig:trajectory} visualizes the predicted trajectory on a QEOxford test sequence, with each scan colour-coded
by localisation outcome: green indicates a correctly localised scan
(translation error $\leq 1$\,m and rotation error $\leq 2^\circ$),
red indicates a failure. UQ-Loc correctly localises 89.1\% of scans at this threshold.
The failures are clustered at a small number of problematic regions.

\subsection{Calibration Analysis}
\label{sec:calibration}

\subsubsection{ECE Metric for 3D Gaussian Predictions.}
For a correctly calibrated heteroscedastic Gaussian, the squared
Mahalanobis residual $r_i^2 = (\cb^*_i - \hat{\cb}_i)^\top \Sig_i^{-1}
(\cb^*_i - \hat{\cb}_i)$ follows a $\chi^2_3$ distribution.
We define the Expected Calibration Error as the mean absolute deviation
between nominal and empirical coverage over a grid of confidence levels
$P = \{0.05, 0.10, \ldots, 0.95\}$:
\begin{equation}
  \text{ECE}
  = \frac{1}{|P|}\sum_{p \in P}
    \left|\hat{E}(p) - p\right|, \qquad
  \hat{E}(p)
  = \frac{1}{N}\sum_{i=1}^{N}
    \mathbf{1}\!\left[r_i^2 \leq F_{\chi^2_3}^{-1}(p)\right],
  \label{eq:ece}
\end{equation}
where $F_{\chi^2_3}^{-1}(p)$ is the $p$-quantile of $\chi^2_3$.
$\hat{E}(p)$ is the empirical coverage at nominal confidence level $p$ — the observed fraction of voxels whose squared Mahalanobis residual falls within the $p$-th quantile of $\chi^2_3$.
A perfectly calibrated model has ECE\,=\,0; empirical coverage
below the diagonal indicates overconfidence, above it underconfidence.

\subsubsection{Results.}
\begin{figure}
  \centering
  \includegraphics[width=0.48\textwidth]{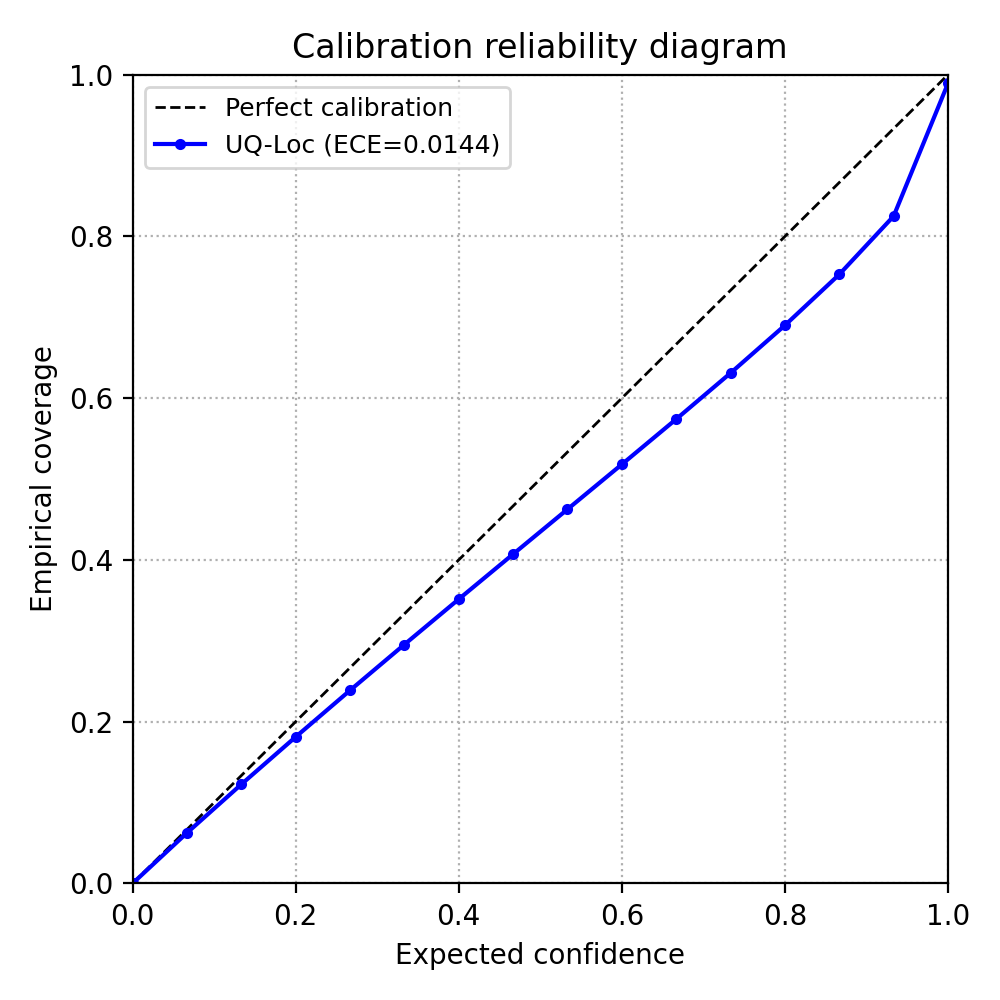}
  \includegraphics[width=0.48\textwidth]{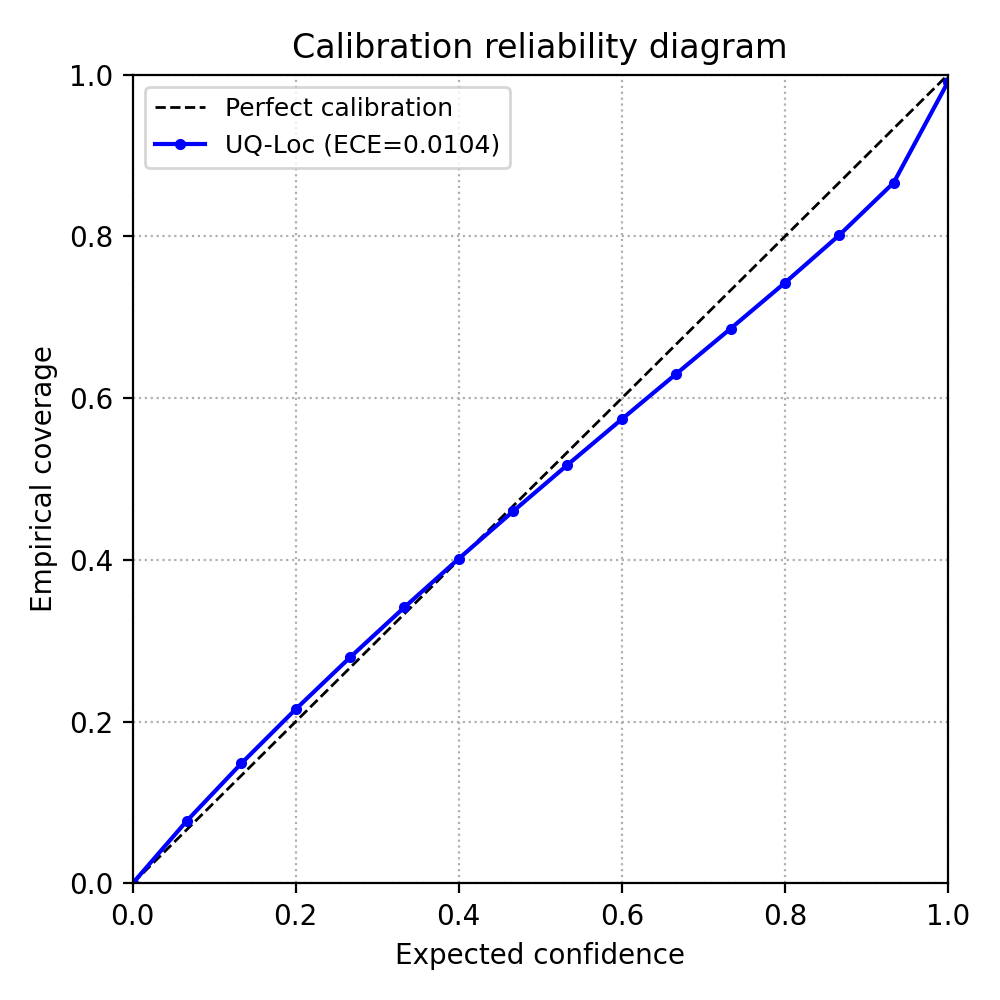}   
  \caption{Reliability diagrams show empirical coverage $\hat{E}(p)$ vs.\ nominal confidence $p$ on the QEOxford (left) and NCLR (right) test sets. The diagonal (dashed) represents perfect calibration.}
  \label{fig:reliability}
\end{figure}

Figure~\ref{fig:reliability} shows the reliability diagram for UQ-Loc
on the test sequences from the QEOxford and NCLT datasets.
The empirical coverage curve closely tracks the perfect-calibration diagonal,
with a slight deviation indicating mild overconfidence.
This is consistent with the mean squared Mahalanobis residual during training exceeding~3 (the expected value of $\chi^2_3$ under perfect calibration).
Nevertheless, deviations remain small (ECE equal to 0.014 on QEOxford and 0.01 on NCLT dataset), confirming that the NLL loss with spatial smoothness produces well-calibrated covariances.



\begin{figure}
  \centering
    \includegraphics[width=1.0\linewidth]{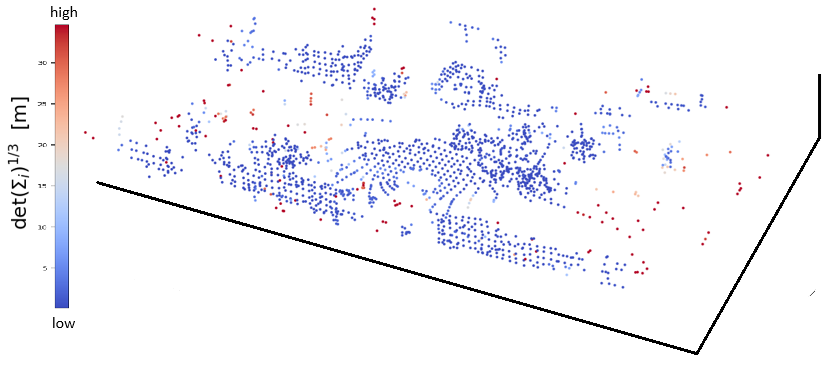}
  \caption{Per-voxel uncertainty magnitude coloured
  blue (low) to red (high) on a scan from QEOxford.}
  \label{fig:qualitative}
\end{figure}

Figure~\ref{fig:qualitative} visualises the predicted uncertainty $\det(\Sigma_i)^{1/3}$ 
for two exemplary scans colour-coded from blue (low) to red (high). 
Oxford RobotCar scans are inherently sparse which limits detailed semantic interpretation of the uncertainty patterns. 
Nevertheless, isolated points at the scene periphery and at greater range 
from the vehicle trajectory tend to be assigned higher uncertainty (redO, which is the expected 
behaviour for a well-calibrated model. 
And the majority of in-range scene points are confidently predicted (blue).

\subsection{Ablation Study}


Table~\ref{tab:ablation} isolates the contribution of each component.
The NLL loss with the Euclidean solver~(B) degrades mean translation
error relative to the L1 baseline~(A) (0.81\,m vs.\ 0.63\,m), confirming
that NLL-trained covariances provide no benefit without a compatible
inlier criterion.
Switching to the Mahalanobis inlier test~(C) recovers accuracy and
improves recall to 48.0\%, demonstrating that the two components are
coupled.
Adding the spatial smoothness regulariser~(D, full model) yields the
best results across all metrics, with recall at the strict $(0.5\,\text{m},\,1^{\circ})$ threshold reaching 49.5\%, confirming that smoother covariance
fields further strengthen the Mahalanobis inlier scoring.
The lower mean TE of~A (0.63\,m vs.\ 0.68\,m) reflects a distribution
shift: NLL training reduces large-error tails, improving median and
recall at the cost of a modest increase in the bulk mean.

\begin{table}
\caption{Ablation study on QEOxford dataset. All models trained
with identical hyperparameters; only the loss, RANSAC inlier test, and smoothness weight ($\lambda$) vary. 
TE/RE are mean translation/rotation errors.}
\label{tab:ablation}
\centering
\setlength{\tabcolsep}{4pt}
\begin{tabular}{clllccccc}
\toprule
\multirow{2}{*}{Cfg}
  & \multirow{2}{*}{Loss}
  & \multirow{2}{*}{$\lambda$}
  & \multirow{2}{*}{RANSAC}
  & \multicolumn{2}{c}{Mean$\downarrow$}
  & \multicolumn{2}{c}{Median$\downarrow$} 
  & {Recall$\uparrow$} \\
\cmidrule(lr){5-6}
\cmidrule(lr){7-8}
  & & &
  & TE\,(m)
  & RE\,(\textdegree) 
  & TE\,(m)
  & RE\,(\textdegree)
  & 0.5m,1\textdegree\\
\midrule
A & L1   & --- & Euclidean  & \textbf{0.63} & 0.83 & 0.46 & 0.72 & 0.428 \\
B & NLL  & 0   & Euclidean & 0.81 & 0.84 & 0.49 & 0.72 & 0.404 \\
C & NLL  & 0  & Mahalanobis     & 0.76 & 0.72  & 0.46 & 0.61 & 0.480 \\
\textbf{D} & NLL  & 0.001 & Mahalanobis   & 0.68 & \textbf{0.70} & \textbf{0.45} & \textbf{0.60} & \textbf{0.495} \\
\bottomrule
\end{tabular}
\end{table}




\section{Conclusion}
\label{sec:conclusion}

We have presented UQ-Loc, which extends LiDAR Scene Coordinate
Regression with per-voxel anisotropic Gaussian covariance estimation.
The Cholesky-parameterised uncertainty head, NLL training loss with
spatial smoothness, and Mahalanobis RANSAC solver each contribute
incrementally to improved pose accuracy, as confirmed by an ablation study.
The first application of ECE to LiDAR SCR
validates that the predicted covariances are well-calibrated and carry
meaningful probabilistic information beyond pose accuracy metrics alone.

Future work includes integration with factor-graph SLAM backends
as uncertainty-weighted constraints, and temporal propagation across sequential frames.

\bibliographystyle{splncs04}
\bibliography{references}

@inproceedings{lightloc,
  title={{LightLoc}: Learning Outdoor LiDAR Localization at Light Speed},
  author={Li, Wen and Liu, Chen and Yu, Shangshu and Liu, Dunqiang and Zhou, Yin and Shen, Siqi and Wen, Chenglu and Wang, Cheng},
  booktitle={Proceedings of the Computer Vision and Pattern Recognition Conference},
  pages={6680--6689},
  year={2025}
}

@inproceedings{sc2pcr,
  title={{SC2-PCR}: a Second Order Spatial Compatibility for Efficient and Robust Point Cloud Registration},
  author={Chen, Zhi and Sun, Kun and Yang, Fan and Tao, Wenbing},
  booktitle={Proceedings of the IEEE/CVF Conference on Computer Vision and Pattern Recognition},
  pages={13221--13231},
  year={2022}
}

@article{dsacstar,
  title={Visual Camera Re-Localization from RGB and RGB-D Images Using DSAC},
  author={Brachmann, Eric and Rother, Carsten},
  journal={IEEE Transactions on Pattern Analysis and Machine Intelligence},
  volume={44},
  number={9},
  pages={5847--5865},
  year={2021},
  publisher={IEEE}
}

@inproceedings{pointnetvlad,
  title={{PointNetVLAD}: Deep Point Cloud Based Retrieval for Large-Scale Place Recognition},
  author={Uy, Mikaela Angelina and Lee, Gim Hee},
  booktitle={Proceedings of the IEEE Conference on Computer Vision and Pattern Recognition},
  pages={4470--4479},
  year={2018}
}

@inproceedings{minkloc3d,
  title={{MinkLoc3D}: Point Cloud Based Large-Scale Place Recognition},
  author={Komorowski, Jacek},
  booktitle={Proceedings of the IEEE/CVF Winter Conference on Applications of Computer Vision},
  pages={1790--1799},
  year={2021}
}

@inproceedings{kendallcipolla,
  title={Geometric Loss Functions for Camera Pose Regression with Deep Learning},
  author={Kendall, Alex and Cipolla, Roberto},
  booktitle={Proceedings of the IEEE Conference on Computer Vision and Pattern Recognition},
  pages={5974--5983},
  year={2017}
}

@inproceedings{he2020deep,
  title={Deep Mixture Density Network for Probabilistic Object Detection},
  author={He, Yihui and Wang, Jianren},
  booktitle={2020 IEEE/RSJ International Conference on Intelligent Robots and Systems (IROS)},
  pages={10550--10555},
  year={2020},
  organization={IEEE}
}

@inproceedings{minkowski,
  title={{4D} Spatio-Temporal {ConvNets}: Minkowski Convolutional Neural Networks},
  author={Choy, Christopher and Gwak, JunYoung and Savarese, Silvio},
  booktitle={Proceedings of the IEEE/CVF Conference on Computer Vision and Pattern Recognition},
  pages={3075--3084},
  year={2019}
}

@article{oxfordrobotcar,
  title={1 Year, 1000 {km}: The Oxford {RobotCar} Dataset},
  author={Maddern, Will and Pascoe, Geoffrey and Linegar, Chris and Newman, Paul},
  journal={The International Journal of Robotics Research},
  volume={36},
  number={1},
  pages={3--15},
  year={2017},
  publisher={SAGE Publications Sage UK: London, England}
}

@article{nclt,
  title={University of Michigan North Campus Long-Term Vision and LiDAR Dataset},
  author={Carlevaris-Bianco, Nicholas and Ushani, Arash K and Eustice, Ryan M},
  journal={The International Journal of Robotics Research},
  volume={35},
  number={9},
  pages={1023--1035},
  year={2016},
  publisher={Sage Publications Sage UK: London, England}
}

@inproceedings{shotton2013,
  author    = {Shotton, Jamie and Glocker, Ben and Zach, Christopher and Izadi, Shahram and Criminisi, Antonio and Fitzgibbon, Andrew},
  title     = {Scene Coordinate Regression Forests for Camera Relocalization in {RGB-D} Images},
  booktitle = {Proceedings of the IEEE/CVF Conference on Computer Vision and Pattern Recognition (CVPR)},
  pages     = {2930--2937},
  year      = {2013}
}

@inproceedings{brachmann2018,
  author    = {Brachmann, Eric and Rother, Carsten},
  title     = {Learning Less is More -- {6D} Camera Localization via {3D} Surface Regression},
  booktitle = {Proceedings of the IEEE/CVF Conference on Computer Vision and Pattern Recognition (CVPR)},
  pages     = {4654--4663},
  year      = {2018}
}

@inproceedings{posenet,
  author    = {Kendall, Alex and Grimes, Matthew and Cipolla, Roberto},
  title     = {{PoseNet}: A Convolutional Network for Real-Time 6-{DoF} Camera Relocalization},
  booktitle = {Proceedings of the IEEE International Conference on
               Computer Vision (ICCV)},
  pages     = {2938--2946},
  year      = {2015}
}

@inproceedings{sgloc,
  title={{SGLoc}: Scene Geometry Encoding for Outdoor {LiDAR} Localization},
  author={Li, Wen and Yu, Shangshu and Wang, Cheng and Hu, Guosheng and Shen, Siqi and Wen, Chenglu},
  booktitle={Proceedings of the IEEE/CVF Conference on Computer Vision and Pattern Recognition},
  pages={9286--9295},
  year={2023}
}

@inproceedings{li2024diffloc,
  title={Diffloc: Diffusion Model for Outdoor {LiDAR} Localization},
  author={Li, Wen and Yang, Yuyang and Yu, Shangshu and Hu, Guosheng and Wen, Chenglu and Cheng, Ming and Wang, Cheng},
  booktitle={Proceedings of the IEEE/CVF Conference on Computer Vision and Pattern Recognition},
  pages={15045--15054},
  year={2024}
}

@article{yu2022lidar,
  title={{LiDAR}-based Localization Using Universal Encoding and Memory-aware Regression},
  author={Yu, Shangshu and Wang, Cheng and Wen, Chenglu and Cheng, Ming and Liu, Minghao and Zhang, Zhihong and Li, Xin},
  journal={Pattern Recognition},
  volume={128},
  pages={108685},
  year={2022},
  publisher={Elsevier}
}

@article{wang2021pointloc,
  title={{PointLoc}: Deep Pose Regressor for Lidar Point Cloud Localization},
  author={Wang, Wei and Wang, Bing and Zhao, Peijun and Chen, Changhao and Clark, Ronald and Yang, Bo and Markham, Andrew and Trigoni, Niki},
  journal={IEEE Sensors Journal},
  volume={22},
  number={1},
  pages={959--968},
  year={2021},
  publisher={IEEE}
}

@article{kendall2017uncertainties,
  title={What uncertainties do we need in bayesian deep learning for computer vision?},
  author={Kendall, Alex and Gal, Yarin},
  journal={Advances in neural information processing systems},
  volume={30},
  year={2017}
}

\end{document}